\documentclass[runninghead]{llncs}
\usepackage[T1]{fontenc}

\usepackage{booktabs}

\usepackage[T1]{fontenc}
\usepackage{xcolor}
\usepackage{hyperref}
\usepackage{graphicx}
\usepackage{amsmath}
\usepackage{listings}
\usepackage{booktabs}
\usepackage{algorithm}
\usepackage{algpseudocode}
\usepackage{array}

\hypersetup{
    colorlinks=true,
    linkcolor=blue,
    urlcolor=cyan
}

\begin{document}
\title{LMV-RPA: Large Model Voting-based Robotic Process Automation}
\author{
    Osama Hosam Abdellaif \and
    Ahmed Ayman \and
    Ali Hamdi
}
\institute{
    \textit{Faculty of Computer Science} \\
    MSA University, Cairo, Egypt \\
    { \email{osama.hosam}, \email{ahmed.ayman26}, \email{ahamdi}}{@msa.edu.eg}
}

%
%

%
%

%
\maketitle              
\begin{abstract}
The need for automating high-volume and unstructured data processing is becoming increasingly critical as organizations aim to improve operational efficiency. In a move to automate workflows, Optical Character Recognition (OCR) plays a key role in it, yet traditional OCR engines frequently struggle with accuracy and efficiency when dealing with complex document layouts and ambiguous text structures. This challenge becomes more pronounced in large-scale tasks, where both speed and precision are essential. In this paper, we address these challenges by introducing LMV-RPA, a new Novel Large Model Voting-based Robotic Process Automation (RPA) system designed to enhance the accuracy and efficiency of OCR workflows. The research problem centers around the limitations of single-engine OCR systems and the need for a more robust solution that would handle complex, high-volume data with greater precision and speed. Our key contribution is the implementing a majority voting mechanism that integrates the outputs from multiple OCR engines—Paddle OCR, Tesseract OCR, Easy OCR, and DocTR—alongside Large Language Models (LLMs), such as LLaMA 3 and Gemini-1.5-pro. This mechanism improves the OCR output to convert it into structured JSON format, significantly enhancing accuracy, particularly for documents with complex and ambiguous layouts. The methodology of this research is the multi-phase pipeline, where each OCR engine’s text extraction is processed by LLMs. The results are then combined using a majority voting mechanism, to make sure that the most accurate text is selected for conversion. This approach not only increase accuracy but also optimizes processing speed, striking a balance between precision and efficiency. LMV-RPA achieves 99\% accuracy in OCR tasks, compared to the baseline model with 94\%, while reducing processing time by 80\%. Benchmark evaluations validate the system’s scalability, demonstrating that LMV-RPA provides a faster, more reliable, and scalable solution for automating large-scale document processing tasks, outperforming existing OCR-RPA integrations.

\end{abstract}


\section{Introduction}

In today’s rapidly evolving digital landscape, businesses want to automate their repetitive and time-consuming tasks to enhance operational efficiency and reduce costs. Robotic Process Automation (RPA) has become one of the most important technologies, enabling organizations to streamline workflows by replicating human interactions with digital systems using software-based robots. These robots can manage tasks such as sending emails, scraping data from websites, handling files, and logging into applications with consistent precision \cite{madakam2019future,sharma2022applications}. Top RPA platforms, including UiPath and Automation Anywhere, have demonstrated significant success in automating structured, rule-based tasks across various industries. However, the integration of Optical Character Recognition (OCR) \cite{al2018arabic,hamdi2021c,al2018enhanced} within these platforms has been essential for automating workflows that involve extracting text from images or scanned documents \cite{tripathi2018learning}. RPA in combination with OCR plays a crucial role in enabling organizations to reallocate human resources from mundane tasks to higher-value work, while also reducing the risk of human error \cite{wewerka2020quantifying,hoffmann2020robotic}.

While these RPA platforms bring great benefits, significant challenges remain when processing unstructured data. Traditional RPA systems \cite{cfb_bots2018difference} excel in automating repetitive, rule-based tasks but often face limitations when dealing with complex workflows that involve unstructured or semi-structured data, such as document text recognition \cite{pekkola2017assessing}. While the UiPath and Automation Anywhere platforms are efficient in many respects, they still face performance bottlenecks, mainly in handling large-scale OCR tasks. The traditional OCR engines frequently employed in these systems, such as Tesseract and Paddle OCR, often struggle with issues related to ambiguous characters, intricate document layouts, and variations  in text structures \cite{tesseract2024ocr,mindee2024doctr}. These challenges limit the effectiveness of traditional RPA platforms \cite{mullakara2024selecting,uipath2024rpa,automation_anywhere2024advanced} in scenarios requiring the processing of high volumes of unstructured data, resulting in inefficiency in both speed and accuracy \cite{ferreira2024evaluation,vialle2020impact}.

To address these limitations, we propose LMV-RPA, a Large Language Model-Driven Robotic Process Automation system that enhances the accuracy and efficiency of OCR workflows. The key contribution of this research is the introduction of a majority voting mechanism between two distinct Large Language Models (LLMs), aimed at improving the conversion of OCR outputs into structured JSON format. Our system integrates multiple OCR engines—Paddle OCR, Tesseract OCR, Easy OCR, and DocTR—into a multi-phase pipeline. Each OCR engine’s output is processed by LLMs, with a majority voting mechanism selecting the most accurate result. This approach allows for increased accuracy, adaptability, and scalability when handling large volumes of unstructured data and outperform common RPA platforms like UiPath and Automation Anywhere \cite{tesseract2024ocr,mindee2024doctr}.

Through extensive benchmarking, we show that LMV-RPA significantly improves both processing speed and accuracy in OCR-driven automation, offering a more scalable and efficient solution compared to existing RPA platforms. This system relieves the limitations of traditional RPA solutions, which are usually unable to handle large-scale, unstructured data processing.

This study introduces several key contributions to the field of OCR-driven automation:
\begin{itemize}
  \item \textbf{Novel LMV-RPA Model:} Developed a robust RPA system that integrates multiple OCR engines and LLMs with a majority voting mechanism, significantly enhancing accuracy (up to 99\%) in text extraction and generation of structured data.
  \item \textbf{Efficiency Improvements:} Showed that LMV-RPA processes tasks 80\% faster than traditional OCR models with only one layer LLM without majority voting while balancing speed and accuracy effectively.
  \item \textbf{Automation Scalability:} Showcased the scalability and operational efficiency of custom RPA systems for complex document workflows, emphasizing the trade-offs between speed and accuracy depending on the requirements of the tasks.
\end{itemize}

The paper is organized as follows. Section \ref{rw} reviews existing research on OCR, RPA, and LLMs, particularly focusing on the challenges faced by traditional platforms such as UiPath and Automation Anywhere. Section \ref{m} introduces the proposed LMV-RPA model, covering directory monitoring, multi-OCR extraction, data structuring via LLMs, majority voting, and continuous automation. Section \ref{rd} presents the results, analyzes their performance, and discusses their implications. Finally, Section \ref{c} summarizes the main contributions of this research and provides a Data and Code Availability section with access to resources used in the study.

\section{Related Work}\label{rw}

The integration of Optical Character Recognition (OCR) with advanced automation frameworks has garnered considerable attention as organizations as they attempt automate the processing of unstructured data. OCR plays a very important role in converting text from scanned documents and images into structured formats; however, traditional OCR engines fail to work on complex text recognition due to inconsistencies in character recognition and layout interpretation \cite{madakam2019future}. To address these challenges, recent research has introduced multi-engine OCR frameworks and Large Language Models (LLMs) to enhance the efficiency and accuracy of text extraction and conversion workflows.

Previous research was dedicated mostly to single-engine OCR solutions, which, while effective for clean and structured text; however, they usually fail to deliver accurate results when faced with noisy images, diverse fonts, or intricate document layouts. For example, Tesseract, a widely adopted OCR engine, struggles with unstructured and ambiguous text data, resulting in frequent recognition errors \cite{tesseract2024ocr}. Such limitations have highlighted the need for more robust systems capable of handling a broader range of input complexity.

To mitigate these limitations, recent studies have started exploring the integration of LLMs to improve OCR output accuracy. LLMs, when paired with OCR engines, have shown remarkable ability in the interpretation of ambiguous characters and correcting structural errors, resulting in enhanced text extraction performance \cite{ferreira2024evaluation}. Additionally, there were some multi-engine approaches have been proposed, combining several OCR engines—such as Tesseract, Paddle OCR, Easy OCR, and DocTR—to process the same input, followed by a majority voting mechanism to determine the most accurate output. These multi-engine frameworks perform better than the traditional single-engine models, particularly when dealing with complex document layouts and ambiguous characters \cite{wewerka2020quantifying,baweja2023comparative}.

Despite these advances, however, there are still some considerable gaps in the literature. Although LLMs have shown huge promise for refining the output of OCR engines, few works have investigated holistic architectures that effectively combine multiple OCR engines with LLMs for large-scale text extraction tasks involving high volumes. Moreover, automation of the process of converting OCR outputs into structured formats, such as JSON-a task quite critical to downstream processing and analysis-remains under-explored. Comparative tests have been done across multi-engine OCR systems are few and far between, further creating a knowledge gap into the scalability and effectiveness of this approach with real-world applications.

This work tries to fill these gaps by proposing a novel framework that will combine several OCR engines with LLMs, and implement a majority vote mechanism in order to enhance the accuracy of the OCR output and automatically convert the extracted text into structured formats such as JSON. We want to show the scalability, accuracy, and efficiency of our proposed solution for large-scale automation driven by OCR through rigorous benchmarks.

\section{Research Methodology}\label{m}
We design the LMV-RPA model to constantly observe a specific directory for any new invoice files, process the images of detected ones using OCR engine\cite{sharma2022applications},  and further refine the extracted data with the help of an LLM in order to create structured JSON \cite{ferreira2024evaluation}. The system then populates the database with the structured data and generates a formatted Template Report automatically; the system works in a completely automated loop\cite{moffitt2018robotic}.

\begin{figure}[!h]
    \includegraphics[width=0.99\textwidth]{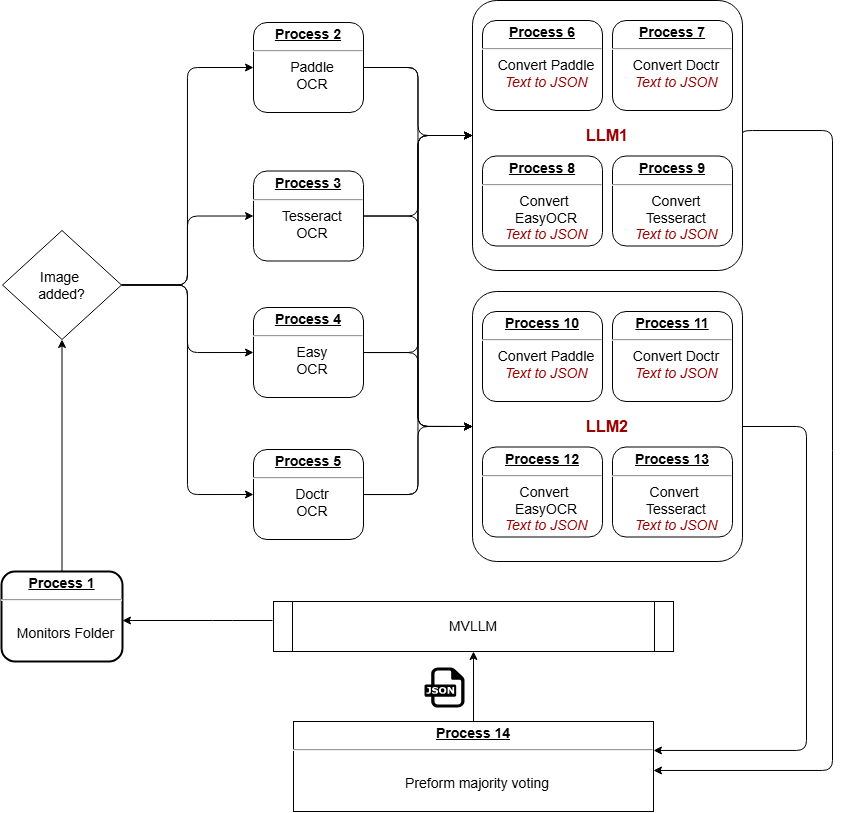} 
    \caption{ System architecture of the proposed LMV-RPA framework. The process starts
with monitoring any particular folder for an image upload (Process 1). Once an image is detected, it is processed by four different OCR engines such as Paddle OCR, Tesseract OCR, Easy OCR, and DocTR OCR in Process 2, Process 3, Process 4, and Process 5, respectively. The outputs from these OCR engines go through two different Large Language Models, LLM1 and LLM2, which convert the extracted text into JSON format. in Processes from 6 to 9 handling the text from Paddle, EasyOCR, Tesseract, and DocTR by LLM1, while the same is done by LLM2 in Processes from 10 through 13. This output undergoes a majority voting mechanism in (Process 14) for selecting the most accurate JSON output to ensure reliable and precise text extraction from complex document layouts. The final structured data is in JSON format.}
    \label{fig1}
\end{figure}

\subsection{LMV-RPA Model}

The LMV-RPA system is designed to continuously monitor a specified directory for new image files, comparing the current files to previously detected ones at regular intervals. Upon detecting a new valid image, it utilizes multiple OCR engines, including PaddleOCR, Tesseract, EasyOCR, and DocTR, to extract text data from the image. This text then undergoes processing via two Large Language Models, which structure the extracted data in JSON format. A majority vote mechanism\cite{smith2023majorityvoting} is used across the JSON outputs for each OCR engine to ensure accuracy. The system will run on a continuous loop, always processing new files as it detects them to make the process automated and efficient for big OCR workflows.

\subsection{Directory Monitoring and File Detection}

The \textit{LMV-RPA} system is designed to continuously monitor a specified directory $\mathcal{D}$ for new image files. At any given time $t$, the system compares the files present in the directory with the files detected at time $t - \Delta t$ (where $\Delta t$ is the monitoring interval). A new file $f_i$ is detected if:
\[
f_i \in \mathcal{F}(t) \setminus \mathcal{F}(t - \Delta t)
\]
If the newly detected file $f_i$ belongs to the set of valid image files $\mathcal{I}$, it is then sent for further processing; otherwise, it is ignored.

\subsection{Multi-OCR Text Extraction}

Once a valid image file $f_i \in \mathcal{I}$ is detected, the \textit{LMV-RPA} system utilizes multiple OCR engines to extract text data $\mathbf{T}_k$. These OCR engines include PaddleOCR, Tesseract, EasyOCR, and DocTR, where each engine is indexed by $k \in \{1, 2, 3, 4\}$. For each OCR engine $\mathcal{O}_k$, the system extracts textual content as:
\[
\mathbf{T}_k = \mathcal{O}_k(f_i)
\]
This multi-engine approach ensures that text data from the image is captured effectively, accounting for variations in document structures and complexities.

\subsection{Data Structuring via Large Language Models (LLMs)}

After extracting the text $\mathbf{T}_k$ from each OCR engine, the raw text is sent to two separate Large Language Models (LLMs) \cite{smith2024gemini,brown2023llama}
for data structuring. Specifically:
\begin{itemize}
    \item The first LLM $\mathcal{L}_1$ processes $\mathbf{T}_k$ to convert it into structured JSON format $\mathbf{J}_k$.
    \item Simultaneously, the second LLM $\mathcal{L}_2$ also processes the same extracted text data $\mathbf{T}_k$.
\end{itemize}

This step is repeated for the outputs from each OCR engine. The transformation process is represented as:
\[
\mathbf{J}_k = \mathcal{L}_1,_2(\mathbf{T}_k)
\]

\subsection{Majority Voting for Final JSON}

To ensure the accuracy of the structured data, the system applies a majority voting mechanism across the different JSON outputs $\mathbf{J}_k$ from the OCR engines. This voting process helps in determining the most accurate and reliable final JSON output:
\[
\mathbf{J}_{final} = \text{MajorityVote}(\mathbf{J}_k)
\]
This ensures that any inconsistencies or errors introduced by individual OCR engines or LLM models are minimized, and the final result is robust and accurate.

\subsection{Continuous Automation}

The \textit{LMV-RPA} system operates in a continuous loop, as shown in \textbf{Algorithm 1}. It monitors the directory for new files and processes them as they are detected. The process continues as long as there are new files to process, ensuring that the system remains operational and automated:
\[
\text{while } t \in \mathcal{T}, \quad \text{perform LMV-RPA process}
\]
\begin{algorithm}[H]
\caption{The LMV-RPA Algorithm}
\label{alg}
\begin{algorithmic}[1]
    \State \textbf{Initialize} monitoring of the directory $\mathcal{D}$
    \While{True}
        \State \textbf{Check} for new files in $\mathcal{D}$
        \If{new file $f_i$ is detected}
            \If{$f_i \in \mathcal{I}$} \Comment{Check if $f_i$ is an image}
                \State \textbf{Extract} text data $\mathbf{T}_k$ using multiple OCR engines $\mathcal{O}_k(f_i)$
                \State \textbf{where} $k \in \{1, 2, 3, 4\}$ for PaddleOCR, Tesseract, EasyOCR, and DocTR
                \For{each OCR engine $\mathcal{O}_k$}
                    \State \textbf{Send} extracted data $\mathbf{T}_k$ to LLM1 $\mathcal{L}1$
                    \State \textbf{Send} extracted data $\mathbf{T}_k$ to LLM2 $\mathcal{L}2$
                    \State \textbf{Transform} $\mathbf{T}_k$ into structured JSON $\mathbf{J}_k$ using LLM $\mathcal{L}1$
                \EndFor
                \State \textbf{Apply majority voting} on the JSON outputs $\mathbf{J}_k$
                \State \textbf{Determine} the most accurate result $\mathbf{J}_{final}$
            \Else
                \State \textbf{Ignore} non-image file
            \EndIf
        \EndIf
        \State \textbf{Continue} monitoring
    \EndWhile
\end{algorithmic}
\end{algorithm}

\section{Benchmark and Experimental Design}
The data used in this research was collected from various sources, primarily focusing on document image datasets that simulate real-world OCR tasks. The utilized dataset consists of 100 diverse document images sourced from platforms like Kaggle and Roboflow \cite{roboflow2024invoice}, as well as Kozłowski’s "Samples of Electronic Invoices" \cite{kozlowski2021samples}. These images were carefully selected to ensure a wide range of text complexity, image quality, and formats to thoroughly test the performance of different OCR engines. The datasets were specifically chosen to evaluate the effectiveness of the LMV-RPA system in handling various document structures, while emphasizing image clarity and its impact on the accuracy and speed of text extraction.

The research was conducted in a simulated document processing environment where the LMV-RPA system was benchmarked using various OCR engines, including Paddle OCR, Tesseract, EasyOCR, and DocTR. All OCR engines were applied to the same dataset to ensure consistency and allow for direct comparison of their performance. The experimental design focused on key metrics such as text extraction speed, accuracy, and the ability to handle complex document structures. Each OCR engine’s output was passed through a Large Language Model (LLM) for JSON conversion, and the final outputs were processed using a majority voting mechanism to determine the most accurate result.

To ensure the reliability of the results, all experiments were conducted on the same hardware and under identical conditions. The LMV-RPA system was evaluated based on overall processing speed, the accuracy of the extracted text, and the efficiency of the LLM in structuring the data. The experimental results were compared against state-of-the-art (SOTA) approaches in OCR-driven automation, focusing on operational efficiency and scalability.

However, there were limitations to the research methodology. Due to reliance on free versions of OCR engines and limited API access, a 5-second delay was implemented between certain operations to ensure compliance with usage restrictions. Additionally, the dataset was limited to 100 images to adhere to daily limits imposed by the data processing tools. These constraints, while necessary for the study, may have impacted the overall performance of some engines. Furthermore, the study was limited to document types relevant to invoice processing, which may not fully reflect the system’s performance on other document types.

\section{Results and Discussion}\label{rd}

Table \ref{tab1} summarizes the performance comparison of our custom LMV-RPA model against the state-of-the-art RPA tools, UiPath and Automation Anywhere across different tasks. The evaluation highlights both the average runtime and accuracy for each model.\newpage

\begin{table}[!htbp]\label{tab1}
\centering
\caption{Comparison of Automation Models by Task with Average}
\label{tab:comparison_with_avg}
\begin{tabular}{lccc}
\toprule
\textbf{Task} & \textbf{Average Run Time} \\
\midrule
\textbf{UiPath} & 212.33 sec \\
\textbf{Automation Anywhere} & 217.8 sec\\
\textbf{LMV-RPA (Ours)} &121.27 sec\\ 
\bottomrule
\end{tabular}
\end{table}

Table \ref{tab1}, we observe a comparison of automation models based on their average run times. UiPath and Automation Anywhere exhibit relatively longer average run times, clocking in at 212.33 seconds and 217.80 seconds, respectively. In contrast, LMV-RPA (Ours) demonstrates a significantly shorter average runtime of 121.27 seconds.

Despite UiPath and Automation Anywhere being well-established platforms with substantial capabilities, LMV-RPA’s shorter runtime showcases its efficiency in processing tasks faster. The reduced runtime of LMV-RPA can be attributed to its optimized architecture, which leverages advanced techniques to maintain high performance while minimizing execution time.

Overall, LMV-RPA stands out as a more efficient solution, offering a significant improvement in processing speed compared to UiPath and Automation Anywhere. This makes it particularly suitable for scenarios where minimizing execution time is critical without sacrificing precision.

\begin{table}[!htbp]\label{tab2}
\centering
\caption{Comparison of Automation Models by Accuracy}
\label{tab:comparison_with_acc}
\begin{tabular}{lcc}
\toprule
\textbf{Models} & \textbf{Accuracy} \\
\midrule
{Traditional Models (1 OCR model + 1 layer LLM)} & 94\% \\
{Our LMV-RPA (4 OCR models + 2 layers LLM with Majority Voting)} & 99\%\\
\bottomrule
\end{tabular}
\end{table}

Table  \ref{tab2} presents a comparison of automation models based on accuracy. The traditional model, which uses 4 OCR engines combined with a single-layer LLM, achieves an accuracy of 94\%. However, the LMV-RPA model (our proposed approach), which incorporates 4 OCR engines and a two-layer LLM with a majority voting mechanism, achieves a significantly higher accuracy of 99\%.

This substantial improvement in accuracy is due to the robustness added by the majority voting mechanism, which cross-verifies the results from two separate LLMs, reducing errors and improving reliability. The comparison highlights the effectiveness of leveraging multiple layers of models and voting mechanisms to enhance the performance of automation models, particularly in tasks where precision is critical.

\section{Conclusion}\label{c}
This study has provided a comprehensive comparison of UiPath, Automation Anywhere, LMV-RPA, focusing on how these three tools are able to execute the automation processes quicker and more accurately. While Both UiPath and Automation Anywhere show regular performance and require longer processing time; the custom-built model of LMV-RPA proved to be very accurate with a result of 99\%, although taking more time in processing tasks. By contrast, LMV-RPA strikes the best balance of speed and accuracy. Thus, it is an extremely effective solution for quick and reliable automation.

These findings underline the potential of custom-built RPA systems in solving particular automation challenges. Indeed, LMV-RPA is best for activities that demand high accuracy, whereas LMV-RPA works best for those tasks that require the highest speed. Hence, it is vital for organizations to consider their needs for automation and balance speed and accuracy requirements while choosing an RPA tool to improve efficiency in operations and meet their business objectives successfully.

\section{Data and Code Availability}
The dataset used in this research is available at the following link: \href{https://www.kaggle.com/datasets/osamahosamabdellatif/high-quality-invoice-i}{data set}. 

\noindent The code repository for this project can be found at: \href{https://github.com/AhmedAyman911/LMV-RPA}{Repo}.

\section*{Acknowledgment}

We extend our heartfelt gratitude to \href{https://msa.edu.eg/msauniversity/}{Modern Science and Arts University (MSA University)} for their funding and continuous support throughout this research. Their generous contribution has been instrumental in the successful completion of our work. We would also like to express our deep thanks to \href{https://aitech.net.au/}{AiTech company} for their invaluable support , guidance throughout and funding this research. Their expertise and assistance have played a crucial role in the development and success of this project.

 \bibliographystyle{bibtex/spmpsci}
 \bibliography{output.bib}

\end{document}